\documentclass{article}
\usepackage{ijcai25}

\usepackage{times}
\usepackage{soul}
\usepackage{url}
\usepackage[hidelinks]{hyperref}
\usepackage[utf8]{inputenc}
\usepackage[small]{caption}
\usepackage{graphicx}
\usepackage{amsmath}
\usepackage{amsthm}
\usepackage{booktabs}
\usepackage{algorithm}
\usepackage{algorithmic}
\usepackage[switch]{lineno}

\usepackage{graphicx}
\usepackage{mathrsfs, bm}
\usepackage{multirow}
\usepackage{array}
\usepackage{appendix}
\usepackage{float}
\usepackage{graphicx}
\usepackage{subfig}
\usepackage{footmisc}
\usepackage{colortbl}
\usepackage{hyperref}
\usepackage{cleveref}
\usepackage{amssymb}

\newtheorem{theorem}{Theorem}

\title{Adaptively Sampling-Reusing-Mixing Decomposed Gradients to Speed Up Sharpness Aware Minimization}

\author{
    Jiaxin Deng, Junbiao Pang
    \affiliations
    Affiliation
    \emails
    email@example.com
}

\begin{document}

\maketitle

\begin{abstract}

Sharpness-Aware Minimization (SAM) improves model generalization but doubles the computational cost of Stochastic Gradient Descent (SGD) by requiring twice the gradient calculations per optimization step. To mitigate this, we propose Adaptively sampling-Reusing-mixing decomposed gradients to significantly accelerate SAM (ARSAM). Concretely, we firstly discover that SAM's gradient can be decomposed into the SGD gradient and the Projection of the Second-order gradient onto the First-order gradient (PSF). Furthermore, we observe that the SGD gradient and PSF dynamically evolve during training, emphasizing the growing role of the PSF to achieve a flat minima.
Therefore, ARSAM is proposed to the reused PSF and the timely updated PSF still maintain the model's generalization ability. Extensive experiments show that ARSAM achieves state-of-the-art accuracies comparable to SAM across diverse network architectures. On CIFAR-10/100, ARSAM is comparable to SAM while providing a speedup of about 40\%. Moreover, ARSAM accelerates optimization for the various challenge tasks (\textit{e.g.}, human pose estimation, and model quantization) without sacrificing performance, demonstrating its broad practicality.% The code is publicly accessible at: \url{https://github.com/ajiaaa/ARSAM}.

\end{abstract}

\section{Introduction}\label{sec:intro}

The powerful generalization ability of Deep Neural Networks (DNNs) has led to significant success in many fields~\cite{chaudhari-2019-Entropy_sgd-IOP}\cite{izmailov-2018-SWA-UAI}. Several studies have verified the relationship between flat minima and the generalization ability~\cite{dinh-2017-sharp_minima-ICML}~\cite{li-2018-visualizing-NIPS}~\cite{jiang-2019-fantastic-ICLR}\cite{liu-2020-loss-NIPS}~\cite{sun-2021-exploring-AAAI}~\cite{Yue2023SALR}. Among these studies, Jiang et al.~\cite{jiang-2019-fantastic-ICLR} explored over 40 complexity measures and demonstrated that a sharpness-based measure exhibits the highest correlation with the generalization.

\begin{figure}[t!]
  \centering
  \includegraphics[width=0.9\linewidth]{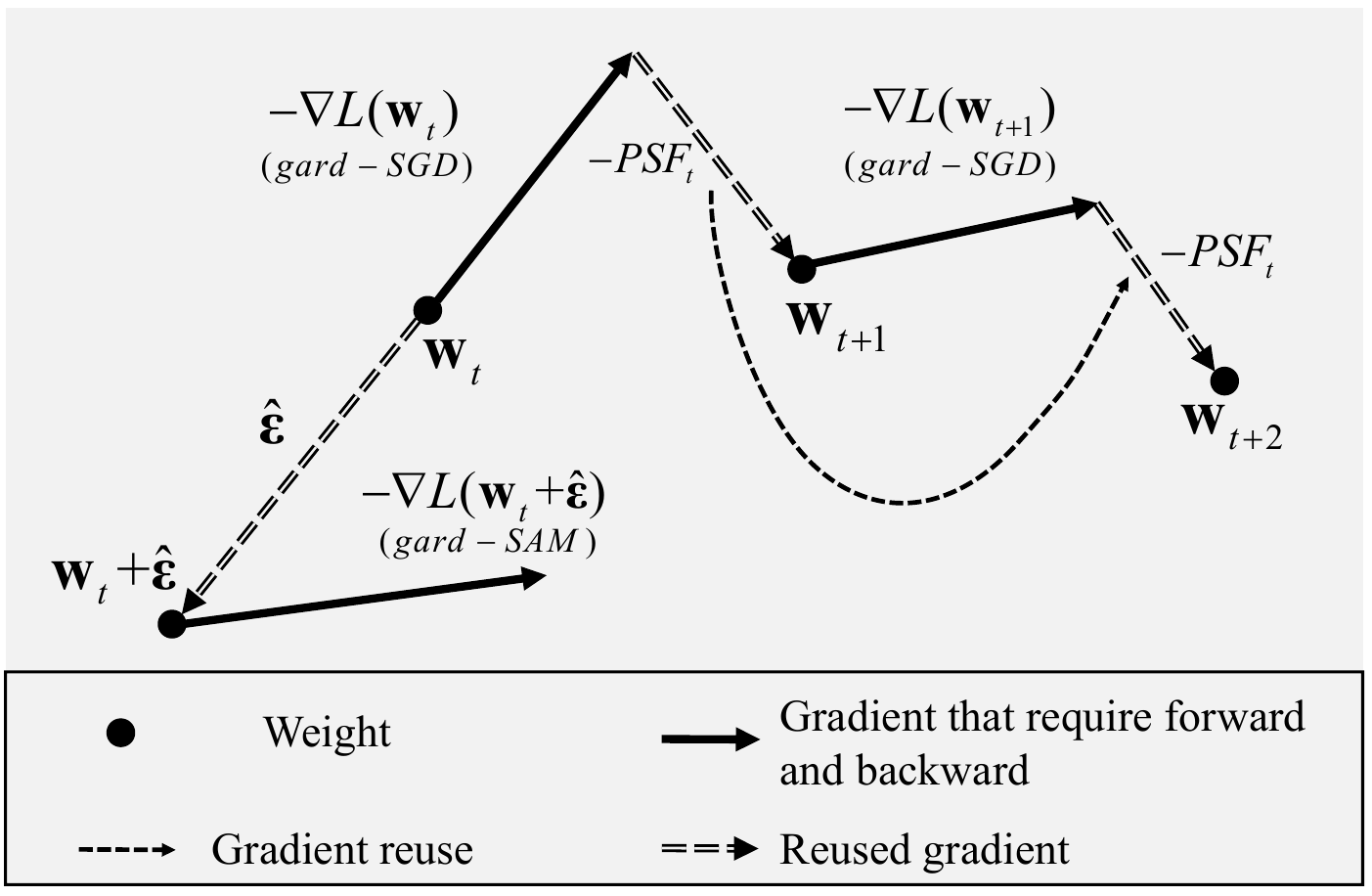}
  \caption{ARSAM speed up SAM by adaptively reusing the decomposed PSF for the next several iterations.
  }
  \label{fig:psf}
\end{figure}

Based on the connection between sharpness of the loss landscape and model generalization, Foret et al. proposed Sharpness Aware Minimization (SAM)~\cite{foret-2020-SAM-ICLR} to find parameter values whose entire neighborhoods have a uniformly low training loss value. Specifically, SAM minimizes both the loss value and the loss sharpness to obtain a flat minimum. SAM and its variants have demonstrated State-Of-The-Art (SOTA) performances across various applications, such as classification~\cite{zhou2023imbsam}\cite{kwon-2021-asam-ICML}, transfer learning ~\cite{du-2022-ESAM-ICLR}\cite{zhuang-2022-GSAM-ICLR}, domain generalization~\cite{dong2024implicit}\cite{xie2024adaptive} and federated learning~\cite{FedGAMMA}.
However, SAM requires two forward and backward passes per iteration, which reduce its optimization speed to half that of SGD. %In some scenarios, costing twice the training time may bring only a marginal improvement (\textit{e.g.}, 0.5\% accuracy) in performances. may not strike an optimal balance between accuracy and efficiency.

\begin{figure*}[!t]
\centering
\subfloat[Resnet-18]{\includegraphics[width=1.7in]{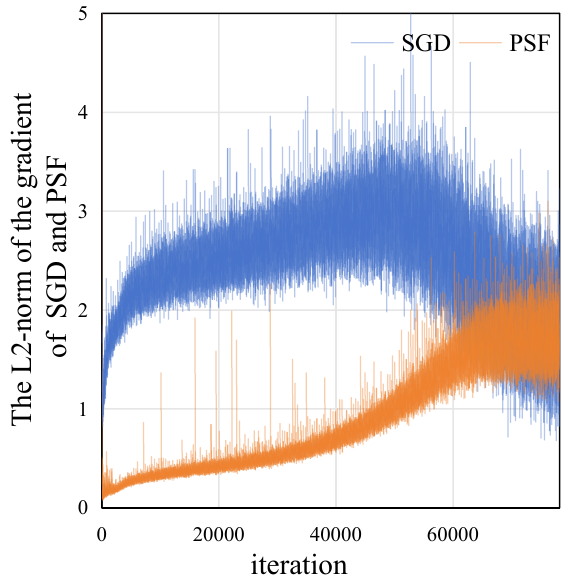}%
\label{fig:fig1a}}
\hfil
\subfloat[WideResNet]{\includegraphics[width=1.7in]{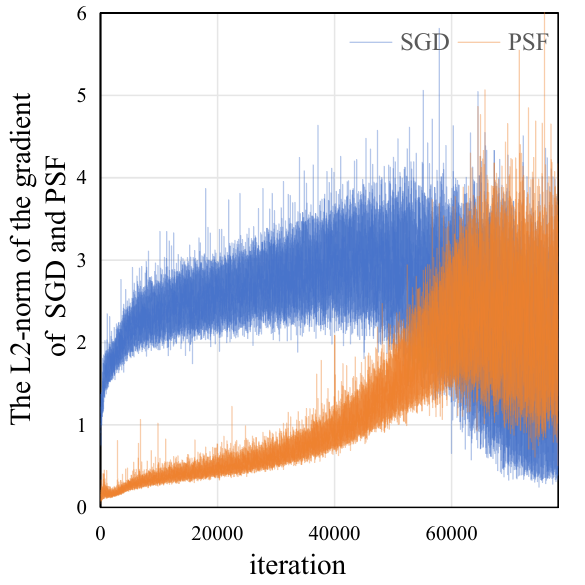}%
\label{fig:fig1b}}
\hfil
\subfloat[PyramidNet]{\includegraphics[width=1.7in]{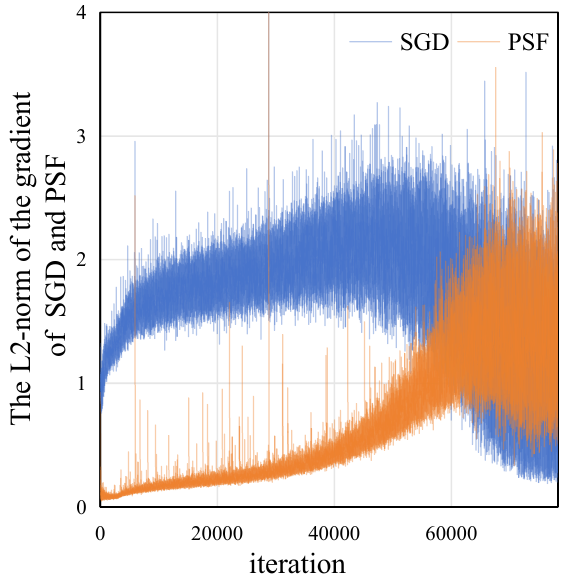}%
\label{fig:fig1c}}
\caption{The trends of $||\nabla L_i^{SGD}||$ and $||\nabla L_i^{PSF}||$ are similar across the different networks (\textit{i.e.}, Resnet-18, WideResNet-28-10, PyramidNet-110) during training (best viewed in color).}
\label{fig:l2norm}
\end{figure*}

In this paper, we observe that SAM's gradient (hereinafter referred to as grad-SAM) can be decomposed into two components: the SGD gradient (hereinafter referred to as grad-SGD) which searches for a local minimum, and the Projection of the Second-order gradient onto the First-order gradient (PSF) which guides SAM toward a flat region. By reusing historical PSF, we approximate the current grad-SAM and thereby accelerate SAM.
Fig.~\ref{fig:psf} illustrates how our method accelerates SAM by reusing the decomposed PSF from the $t$ iteration to update the weights at the $t+1$ iteration. 
Therefore, only one forward and one backward are needed to update the weights $\mathbf{w}_{t+1}$, thus improving the optimization speed.
% fig这里说的不清楚，应该说明1）分解，2）加速的原理和直觉

Importantly, we trained Resnet-18, WideResNet-28-10 and PyramidNet-110 on the CIFAR-100 by SAM and recorded the changes in the L2-norm of the SGD (hereinafter referred to as L2-SGD) (\textit{i.e.}, $||\nabla L_i^{SGD}||$) and that of the the L2-norm of PSF (hereinafter referred to as L2-PSF) (\textit{i.e.}, $||\nabla L_i^{PSF}||$) at the $i$-th iteration in Fig.\ref{fig:l2norm}. Two observations can be drawn from Fig.\ref{fig:l2norm} as follows:
\begin{itemize}
\item Both L2-SGD and L2-PSF change dynamically during training, indicating that the contributions of grad-SGD and PSF dynamically evolve throughout the learning process. In contrast, AE-SAM~\cite{jiang-2022-aesam-ICLR} only considers changes in L2-SGD during training. 
\item L2-PSF increases faster than L2-SGD with the increase of the iteration, especially at the later training stage. This indicates that PSF plays a key role in achieving a flat loss landscape at the later stage of training.
\end{itemize}

Inspired by the above observations, we propose accelerating SAM by adaptively reusing the decomposed PSF based on the contribution of the PSF. We utilize the ratio between the L2-PSF and L2-SGD as an indicator to reflect the contribution of the PSF during training. 

The other important question is how to sample which iteration is used to decompose grad-SAM into the grad-SGD and the PSF. We built an autoregressive model based on the proposed indicator to adaptively adjust the sampling frequency. Specially, when the contribution of the PSF decreases, the autoregressive model reduces the computation frequency and reuses the historical PSF; otherwise, the autoregressive model increases the computation frequency of the PSF to improve optimization accuracy. Compared to uniform sampling (\textit{e.g.}, SAM-5), non-uniform sampling highlights the role of the PSF at the different training stages. Unlike AE-SAM, we consider the different roles of PSF and grad-SGD at each iteration. The mixture of the reused PSF and the timely updated PSF would maintain the model's generalization ability. 
% 给出采样的自回归的依据和有点，还有比baseline的优点。写在后面了

In summary, our contributions are as follows:
\begin{itemize}
    \item To the best of our knowledge, we firstly uncover that the grad-SAM can be decomposed into the grad-SGD and PSF. The discovery suggests that the gradient norm~\cite{zhao-2022-penalizing-ICML} is equivalent to SAM, except for the detailed optimization rule. 
    \item We also observe that the L2-PSF increases gradually during training across different networks, motivating us to adaptively compute the PSF to accelerate SAM. Thus, we introduce ARSAM, which adaptively reuses the decomposed PSF to not only preserve the model's generalization ability but also speed up SAM.
    \item The empirical results demonstrate that ARSAM achieves a 40\% acceleration in speed compared to SAM while ensuring almost no decrease in model generalization ability. Additionally, ARSAM can be applied to various tasks, such as human pose estimation, and model quantization.
\end{itemize}

\section{Related Works}
\label{sec:related}
\subsection{Background of SAM}
Foret et al.~\cite{foret-2020-SAM-ICLR} introduced an effective approach to improving the model's generalization ability. The optimization process of SAM can be viewed as addressing a minimax optimization problem as follows:
\begin{equation}\label{equ:sam}
\begin{aligned}
\mathop {\min } \limits_\mathbf{w} & \; {L^{SAM}}(\mathbf{w}) + \lambda ||\mathbf{w}||_2^2 \\
& where \; {L^{SAM}}(\mathbf{w}) = \mathop {\max }\limits_{||\bm{\varepsilon} || \le \rho } L(\mathbf{w} + \bm{\varepsilon} ),
\end{aligned}
\end{equation}
where $\mathbf{w}$ represents the weights of the network, $\bm{\varepsilon}$ represents the perturbation of weights $\mathbf{w}$ in a Euclidean ball with the radius $\rho$ $(\rho>0)$, $L(\cdot)$ is the loss function, and $\lambda ||\mathbf{w}||_2^2$ is a standard L2 regularization term. 

SAM utilizes Taylor expansion to search for the maximum perturbed loss ($L^{SAM}(\mathbf{w})$) in local parameter space as follows:
\begin{equation}\label{equ:psf}
\begin{aligned}
\mathop {\arg \max }\limits_{||\bm{\varepsilon} || \le \rho } \; L(\mathbf{w} + \bm{\varepsilon} ) & \approx \mathop {\arg \max }\limits_{||\bm{\varepsilon} || \le \rho } \; L(\mathbf{w}) + {\bm{\varepsilon} ^{\top}}{\nabla _\mathbf{w}}L(\mathbf{w}) \\
&= \mathop {\arg \max }\limits_{||\bm{\varepsilon} || \le \rho } \; {\bm{\varepsilon} ^{\top}}{\nabla _\mathbf{w}}L(\mathbf{w}).
\end{aligned}
\end{equation}
By solving Eq.~\eqref{equ:psf}, SAM obtains the perturbation as follows:
\begin{equation}\label{equ:e_max}
\begin{aligned}
\hat{ \bm{\varepsilon} } = \rho {\nabla _\mathbf{w}}L(\mathbf{w})/||{\nabla _\mathbf{w}}L(\mathbf{w})||.
\end{aligned}
\end{equation}
$\hat{ \bm{\varepsilon} }$ can maximize the perturbed loss. SAM computes the gradient $\nabla _\mathbf{w}L(\mathbf{w})$ to obtain the perturbation. Substituting the perturbation $\hat{\bm{\varepsilon} }$ back into Eq.~\eqref{equ:sam} and differentiating, we then have:
\begin{equation}\label{equ:gs}
\begin{aligned}
{\nabla _{\mathbf{w}}}&{{{L}}^{SAM}}({\mathbf{w}}) \approx {\nabla _{\mathbf{w}}}L({\mathbf{w}} + \hat {\bm{\varepsilon}} ({\mathbf{w}})) \\ &= \frac{{d({\mathbf{w}} + \hat {\bm{\varepsilon}} ({\mathbf{w}}))}}{{d{\mathbf{w}}}}{\nabla _{\mathbf{w}}}L({\mathbf{w}}){|_{{\mathbf{w}} + \hat {\bm{\varepsilon}} ({\mathbf{w}})}}\\
 &= {\nabla _{\mathbf{w}}}L({\mathbf{w}}){|_{{\mathbf{w}} + \hat {\bm{\varepsilon}} ({\mathbf{w}})}} + \frac{{d\hat {\bm{\varepsilon}} ({\mathbf{w}})}}{{d{\mathbf{w}}}}{\nabla _{\mathbf{w}}}L({\mathbf{w}}){|_{{\mathbf{w}} + \hat {\bm{\varepsilon}} ({\mathbf{w}})}}.
\end{aligned}
\end{equation}
By dropping the second-order terms in Eq.\eqref{equ:gs}, SAM calculates the gradient at $\mathbf{w}+\bm{\hat\varepsilon}$ as follows:
\begin{equation}\label{equ:grad_sam}
\begin{aligned}
{\nabla _{\mathbf{w}}}&{{{L}}^{SAM}}({\mathbf{w}}) \approx {\nabla _\mathbf{w}}L(\mathbf{w}){|_{\mathbf{w} + \bm{\hat\varepsilon} }}.
\end{aligned}
\end{equation}
Finally, SAM computes gradient at the perturbed point ${\nabla_\mathbf{w}}L(\mathbf{w}){|_{\mathbf{w} + \bm{\hat\varepsilon}}}$ for optimization. From the above process, we observe that SAM necessitates two forward and backward operations to update weights once.

\subsection{Efficient optimization methods for SAM}
Methods for speeding up SAM are broadly categorized into two groups: 1) accelerating the gradient computation process and 2) reducing gradient computation.

In the first category, Du et al. propose SAF and MESA~\cite{du-2022-saf-NIPS}, which employ trajectory loss to approximate sharpness through a surrogate sharpness measure. However, SAF requires significant memory to store all output results, while MESA needs to maintain an exponential moving average model during training. Du et al. propose ESAM~\cite{du-2022-ESAM-ICLR}, which employs Stochastic Weight Perturbation (SWP) and Sharpness Sensitive Data Selection (SDS) strategy to reduce computation.
SWP approximates weight perturbation using a subset of model weights, while SDS focuses on training data that most impact sharpness. Deng et al. propose AUSAM~\cite{Deng2024AsymptoticUS} to sample a subset of data from the mini-batch based on the estimated average gradient norm, reducing the forward and backward time.

In the second category, Liu et al. propose LookSAM~\cite{liu-2022-looksam-CVPR} which periodically calculates grad-SAM every 5 iterations and reuses previous gradients in other iterations.
However, this uniform gradient sampling fails to reflect the importance of sharpness optimization in the overall optimization process. Subsequently, Jiang et al. claims that the SAM update is more useful in sharp regions than in flat regions. As a result, they designs AE-SAM~\cite{jiang-2022-aesam-ICLR} to adaptively employ SAM based on the loss landscape geometry. ARSAM belongs to this research line. Unlike the approaches mentioned above, ARSAM adaptively computes grad-SAM to accelerate optimization and reuses PSF to ensure the model's generalization performance.

\section{Method}
\subsection{Gradient Composition of SAM}

We rewrite Eq.~\eqref{equ:grad_sam} by Taylor expansion and substitute $\bm{\hat \varepsilon}$ to obtain the following:
\begin{equation}\label{equ:method_taylor_expansion}
\begin{aligned}
{\nabla _\mathbf{w}}L(\mathbf{w} + \bm{\hat\varepsilon} ) &\approx {\nabla _\mathbf{w}}(L(\mathbf{w}) + \bm{\hat \varepsilon} {\nabla _\mathbf{w}}L(\mathbf{w}))\\
 &= {\nabla _\mathbf{w}}(L(\mathbf{w}) + \rho ||{\nabla _\mathbf{w}}L(\mathbf{w}){\rm{||)}}.
\end{aligned}
\end{equation}
Eq.~\eqref{equ:method_taylor_expansion} indicates that the grad-SAM can be considered as a combination of the grad-SGD $\nabla _\mathbf{w} L(\mathbf{w})$, and the gradient of the L2-norm of SGD's gradient $\nabla _\mathbf{w} \rho ||{\nabla _\mathbf{w}}L(\mathbf{w})|| $.

We expand the second term in Eq.~\eqref{equ:method_taylor_expansion} as follows:
\begin{equation}\label{equ:method_exp_grad}
{\nabla _\mathbf{w}}(\rho ||{\nabla _\mathbf{w}}L(\mathbf{w}){\rm{||)}} = \rho \frac{{{\nabla _\mathbf{w}}L(\mathbf{w}) \nabla _\mathbf{w}^2L(\mathbf{w})}}{{||{\nabla _\mathbf{w}}L(\mathbf{w}){\rm{||}}}}.
\end{equation}
The gradient of the L2-norm of SGD's gradient (hereinafter referred to as grad-L2-SGD) can be transformed into the Projection of the Second-order gradient onto the First-order gradient, which we refer to as PSF. Eq.~\eqref{equ:method_taylor_expansion} and Eq.~\eqref{equ:method_exp_grad} indicate that SAM's gradient can be decomposed into the SGD gradient and the PSF. Naturally, the PSF prompts SAM to find a flat minima. 

\textbf{Connect PSF to gradient norm~\cite{zhao-2022-penalizing-ICML}:} By comparing Eq.(6) in \cite{zhao-2022-penalizing-ICML} ($\nabla_{\theta} L(\theta) = \nabla_{\theta} L_{S}(\theta) + \lambda \cdot \nabla_{\theta}^2 L_{S}(\theta) \frac{\nabla_{\theta} L_{S}(\theta)}{||\nabla_{\theta} L_{S}(\theta)||}$) with Eq.~\eqref{equ:method_exp_grad} in our paper, we discover that the PSF ($\frac{{{\nabla _\mathbf{w}}L(\mathbf{w}) \nabla _\mathbf{w}^2L(\mathbf{w})}}{{||{\nabla _\mathbf{w}}L(\mathbf{w}){\rm{||}}}}$) is consistent with the second term on the right side of Eq. (6) in \cite{zhao-2022-penalizing-ICML}. This indicates that our proposed acceleration method is applicable not only to SAM but also to the gradient norm \cite{zhao-2022-penalizing-ICML}.

% 说明和SAM的关系，说明我们的方法。

\begin{theorem} \label{lemma1}
Let $\nabla _\mathbf{w}^2L(\mathbf{w})$ be a hessian matrix with n eigenvalues, then the L2-PSF has an upper bound as follows:
\begin{equation}\label{equ:method_lemma}
\left\| {{\nabla _\mathbf{w}}(\rho \left\| {{\nabla _\mathbf{w}}L(\mathbf{w})} \right\|)} \right\| \le \rho \sum\limits_{i = 1}^n {{\delta _i}}, 
\end{equation}
where $\delta _i$ is the $i$-th eigenvalue.
\end{theorem}

\textbf{Remark.} Theorem~\ref{lemma1} discovers that the L2-PSF is related to the eigenvalues of the matrix $\nabla _\mathbf{w}^2L(\mathbf{w})$. As the sum of the eigenvalues decreases, the L2-PSF becomes smaller.

Without considering the direction of the PSF, we empirically studied the behavior of the  L2-PSF in Fig.~\ref{fig:l2norm} and observed that the L2-PSF (\textit{i.e.}, $||\nabla L_i^{PSF}||=||{\nabla L_i^{SAM}-\nabla L_i^{SGD}}||$) gradually increases during training. This phenomenon reflects that the significance of the PSF dynamically changes during the training process. One natural question is: When and how to adaptively sample and reuse the PSF based on its behavior to accelerate optimization while guaranteeing generalization?

\subsection{Sampling-Reusing-Mixing Decomposed Gradient}

\subsubsection{Sampling the PSF based on the ratio}

Firstly, we use the ratio of L2-PSF to L2-SGD as an indicator to evaluate the contribution of PSF in finding a local minimum, as follows:
\begin{equation}\label{equ:method_r}
\begin{aligned}
c_{i} = \frac{||\nabla L_{i}^{PSF}||}{||\nabla L_{i}^{SGD}||},
\end{aligned}
\end{equation}
where $||\nabla L_{i}^{PSF}||$ and $||\nabla L_{i}^{SGD}||$ are the L2-PSF and the L2-SGD at the $i$-th iteration, respectively. We only considered the magnitude of the gradients, not their directions, as Theorem~\ref{lemma1} discovers that the magnitude is enough to calibrate the contribution of the PSF and SGD during the optimization. 

We have observed that the distribution of the $c_i$ at the same iteration from the different sizes and architecture of networks is very different, as illustrated in Fig.~\ref{fig:l2norm}. If we directly use $c_i$ in Eq.~\eqref{equ:method_r}, it becomes difficult to keep its range under control across different networks. Therefore, we consider the relative change of the indicator $c_i$ at the current iteration, which is normalized by the previous iteration as follows:
\begin{align}
r_i &= \frac{c_i-c_{i-1}}{c_{i-1}}. \label{equ:method_c_norm_i}
\end{align}
The aim of Eq.~\eqref{equ:method_c_norm_i} is to model the change in $c_i$ through $c_i - c_{i-1}$ and normalize the magnitude of $c_i$ across different models and datasets using $1/c_i$. The normalization in~\eqref{equ:method_c_norm_i} helps different networks have the same characteristic as illustrated in Fig.~\ref{fig:l2norm}.

Furthermore, to smooth $r_{i}$, we apply the Exponential Moving Average (EMA) strategy to smooth $r_i$ as follows:
\begin{align}
\hat{r}_i &= \beta \cdot \hat{r}_{i-1} + (1-\beta) \cdot r_i,
\end{align}
where $\beta$ is the weighted weight, typically set to 0.9. 

We construct an autoregressive model to transform the smoothed $\hat{r}_i$ into the number of PSF to be sampled. We chose the autoregressive model because of its simplicity and generality; in addition, it is easy to implement and does not impose a heavy computational burden. We divide the total $N$ iterations into a series of segments where each length is $M$. In other words, for every $M$ iteration, we update the number of times the PSF will be sampled in the next segment as follows:
\begin{equation}\label{equ:method_s}
s_{j+1}= s_{j} \cdot (1 + \alpha \cdot \hat{r}_{k}),
\end{equation}
where $s_{j+1}$ represent that how many PSF would be sampled in the next $M$ iterations, $k = j \cdot M$ represent the index of iteration in a training, the hyperparameter $\alpha$ adjusts the increase/decrease ratio of $s_{j+1}$. We set $s_0$ to $1$ in our experiments. That is, only 1 PFS is computed in the first segment. Finally, the sampling probability of PSF for the next $M$ iterations is defined as follows:
\begin{equation}\label{equ:method_p}
p_{j+1} = \frac{s_{j+1}}{M}.
\end{equation} 
where $ j=1,\ldots, M$. We perform gradient sampling with the probability $p_{j+1}$ for the next $1,\ldots, M$ iterations.

\textbf{Approximate the L2-PSF and L2-SGD:} As the model size increases, the computational cost of the L2-norm grows. As shown in Appendix 1.5, we observe that the L2-SGD and L2-PSF in the last few layers of the model also capture the changes in the gradient of SGD and PSF. % 给出图的证据给出比值的证据
Thus, we replace $||\nabla L^{SGD}||$ and $||\nabla L^{PSF}||$ with the L2-SGD and the L2-PSF in the last few layers of the model, respectively. 

\textbf{Advantages over LookSAM and AE-SAM:} In contrast to updating grad-SAM uniformly in LookSAM~\cite{liu-2022-looksam-CVPR}, ARSAM computes the PSF in a non-uniform manner to reflect its importance during training. This approach helps schedule the frequency of PSF computation more effectively, thus maintaining the model’s generalization ability. 

In contrast to AE-SAM, which establishes a distribution about L2-SGD to decide whether to replace SAM with SGD, ARSAM uses the ratio between L2-PSF and L2-SGD (\textit{i.e.} Eq.~\eqref{equ:method_r}) as an indicator to reuse PSF. If the reuse error is small enough, it can be treated as minor noise in the gradient. As the \cite{keskar-2016-large_batch-ICLR} explains, this noise helps push the iterates away from sharp minimizers and toward flatter ones, potentially allowing ARSAM to find a flatter minimum.% 说明对近似的理解
\begin{algorithm}[tb]
    %\begin{small}
    \caption{{Pseudocode of the proposed ARSAM}}
    {\bf Require:}
    The training dataset, the learning rate $\eta$, parameters $\alpha$,  $\rho$, $M$, $s_1$ and $I_{start}$.
    \begin{algorithmic}[1]
    \FOR{$i = 1,2,\cdot\cdot\cdot$}
    \STATE Calculate the grad-SGD $\nabla L^{SGD}_i$;
    \IF{$i <= I_{start}$ \OR sampling}
    \STATE Calculate the grad-SAM $\nabla L^{SAM}_i$ in Eq.~\eqref{equ:grad_sam};
    \STATE Calculate the PSF $\nabla L^{PSF}_i$;
    \STATE Update the weights according to Eq.~\eqref{equ:sampling_opt};
    \ELSE
    \STATE Update the weights according to Eq.~\eqref{equ:non-sampling_opt};
    \ENDIF
    \IF{$i > I_{start}$ \AND $i$ \% $M = 0$}
    \STATE Update the sampling probability by Eq.~\eqref{equ:method_p};
    \ENDIF
    \ENDFOR
    \end{algorithmic}
    \label{algorithm:1}
   % \end{small}
\end{algorithm}

\subsubsection{Reusing-Mixing decomposed gradients} \label{sec:reuse}
When the computation probability of PSF is obtained, we determine whether to compute PSF in each iteration based on this probability. The weight update process is as follows:
\begin{itemize}
    \item When computing the PSF, the optimization rule can be written as follows:
\begin{equation}\label{equ:sampling_opt}
\centering
\begin{aligned}
{\mathbf{w}_{i+1}} = {\mathbf{w}_{i}} - \eta \nabla L_i^{SAM},
\end{aligned}
\end{equation}
where $\eta$ is the learning rate. Eq.~\eqref{equ:sampling_opt} is essentially the optimization rule for SAM.
    \item When PSF is not computed, we reuse the PSF from the last computed iteration, the optimization rule can be written as follows:
\begin{equation}\label{equ:non-sampling_opt}
{\mathbf{w}_{i+1}} = {\mathbf{w}_{i}} - \eta (\nabla L_i^{SGD} +  \nabla L_{\hat{i}}^{PSF}),
\end{equation}
where $\nabla L_{\hat{i}}^{PSF}$ is the PSF at the last computed iteration.
\end{itemize}
Algorithm \ref{algorithm:1} shows the overall proposed algorithm. 

\subsubsection{Convergence analysis}
We further analyze the convergence properties of ARSAM as follows.
\begin{theorem} \label{lemma2}
Suppose $L^{SGD}(\mathbf{w})$ is $\tau$-Lipschitz smooth and $|L^{SGD}(\mathbf{w})|$ is bounded by $U$. For any $t \in \{0,...,T\}$ and any $\mathbf{w} \in W$, suppose we can obtain bounded observations as follows:
\begin{equation}\label{equ:bound_obs}
\centering
\begin{aligned}
&||\hat{\bm{\xi}}_t||^2 \le G_1, \; \mathbb{E}[\nabla L_t^{SGD}(\mathbf{w})]=\mathbb{E}[\hat{\mathbf{g}}_t]= \mathbf{g}_t, \; \|\hat{\mathbf{g}}_t\|\leq G_2, \\
&\mathbb{E}[\nabla L_t^{PSF}(\mathbf{w})]=\mathbb{E}[\hat{\mathbf{h}}_t]= \mathbf{h}_t, \; \|\hat{\mathbf{h}}_t\|\leq G_3. 
\end{aligned}
\end{equation}
Then with learning rate $\eta_t = \frac{\eta_0}{\sqrt{t}}$, we have the following bound for ARSAM:
\begin{equation}\label{equ:convergence}
\centering
\begin{aligned}
\frac{1}{T}\sum\limits_{t=1}^T &{\mathbb{E}[||\mathbf{g}_t + \mathbf{h}_t + \bm{\xi}_t||^2]} \le \frac{{2D_1} + {2D_2}InT}{\sqrt{T}} +2\tau,
\end{aligned}
\end{equation}
where $\bm{\xi}_t$ can be specifically defined as follows:
\begin{equation}
\tiny
\begin{aligned}
\boldsymbol{\xi }_t=\left\{ \begin{array}{l}
	-\mathbf{{h}}_t\ \ \ \ \ When\ updating\ with\ the\ gradient\ of\ SGD\\
	0\ \ \ \ \ \ \ \ When\ updating\ with\ gradient\ reusing\\
	\mathbf{\hat{h}}_t-\mathbf{h}_t\ \ When\ updating\ with\ the\ gradient\ of\ SAM\\
\end{array} \right. 
\end{aligned}
\end{equation}
where the true PSF is $\mathbf{h}_t$ and the reused PSF is $\mathbf{\hat{h}}_t$.
\end{theorem}
\textbf{Remark.} Theorem~\ref{lemma2} uncovers that the convergence of ARSAM is affected by $D1$, $D2$, $T$, and $\tau$. Concretely, $D1$ and $D2$ are dependent on factors such as $\tau$, $\eta_0$, $G_1$, $G_2$, $G_3$ and $U$, all of which must be controlled within a certain range. Therefore, the bounded gradients mean that the optimizer would converge to a local minimum, as discussed in \cite{andriushchenko-2022-towards_understanding-ICML}.

\subsubsection{The difference between the reuse PSF and the true PSF}

\begin{theorem} \label{lemma3}
Assume that the weight at $t$-th iteration is $\mathbf{w}_t$, and the index of the reused gradients is $t+n (n>0)$, \textit{i.e.}, $\mathbf{w}_{t+n}$. Assuming that $\frac{{\nabla {L_B}({\mathbf{w}_{t}})}}{{\left\| {\nabla {L_B}({\mathbf{w}_{t}})} \right\|}}$ is Lipschitz continuous and $\tau$ is Lipschitz constant. The expected error between the reused PSF with the real PSF is bounded as follows:
\begin{equation}\label{equ:error_reused}
\centering
\begin{small}
\begin{aligned}
&\left\| {{\mathbb{E}_B}\left[{\nabla ^2}{L_B}({\mathbf{w}_{t + n}})\frac{{\nabla {L_B}({\mathbf{w}_{t + n}})}}{{\left\| {\nabla {L_B}({\mathbf{w}_{t + n}})} \right\|}} - {\nabla ^2}{L_B}({\mathbf{w}_t})\frac{{\nabla {L_B}({\mathbf{w}_t})}}{{\left\| {\nabla {L_B}({\mathbf{w}_t})} \right\|}}\right]} \right\| \\
&\quad\quad \le n\eta \tau \mathbb{E}\left[Tr({\nabla ^2}L({\mathbf{w}_t}))\right] \left\| \mathbb{E}[\nabla L({\mathbf{w}})] \right\|
\end{aligned}
\end{small}
\end{equation}
where $B$ represents a mini-batch data.
\end{theorem}

\textbf{Remark.} The Lipschitz constant and the trace are associated with the flatness of the loss landscape. Theorem \ref{lemma3} shows that as the optimizer reaches a flat loss landscape, the upper bound in Eq.~\eqref{equ:error_reused} would become small. Moreover, we could reduce the upper bound of the error by controlling the distance between $\mathbf{w}_{t}$ and $\mathbf{w}_{t+n}$. Therefore, the gradient reuse in Section \ref{sec:reuse} is empirically applied at the two consecutive iterations after sampling iteration. 

\textbf{Acceleration ratio of ARSAM over SAM}
As shown in Appendix 1.4, we discover that the mean of the indicator $\hat{r}_i$ gradually increases with the index of iterations and approximately follows a quadratic function.
To simplify and formalize the acceleration of ARSAM, we assume that the average trend of the indicator $\hat{r}_i$ follows the function $\hat{r}_i = \gamma \cdot i^2$, where $i$ represents the iteration index, and $\gamma$ reflects the curvature of the function. The total number of times ARSAM samples the PSF during the entire training process can be expressed as:
\begin{equation}\label{app-eq19}
\begin{aligned}
s^* = \sum\limits_{j = 1}^{I/M} {{s_j}} , \quad \textit{where}\; {s_j} = {s_{j - 1}}(1 + \alpha \cdot \gamma \cdot (j \cdot M)^2),
\end{aligned}
\end{equation}
where $\alpha$ controls the variation of $n_{j}$, and $M$ indicates that $\hat{r}_i$ is sampled every $M$ intervals, $I$ represents the total number of iterations. Assume that gradient computation dominates the optimization time, with other factors neglected, and that each gradient computation takes the same amount of time. The optimization speed of ARSAM compared to SAM is approximately as follows:
\begin{equation}\label{app-eq20}
\begin{aligned}
v = {2I}/{(I+s^*)}.
\end{aligned}
\end{equation}
These demonstrate that ARSAM effectively accelerates SAM. More details can be found in Appendix 1.4.

\section{Experimental Results}
\subsection{Setup}
\subsubsection{Datasets and Models}

To verify the effectiveness of ARSAM, we conduct experiments on CIFAR-10, CIFAR-100 \cite{krizhevsky2009learning} and Tiny-ImageNet \cite{le-2015-tiny} image classification benchmark datasets. 
%Tiny-ImageNet is a subset of the larger ImageNet dataset, which is also a popular image dataset widely used for machine learning and computer vision tasks.
We employ a variety of architectures to evaluate the performance and training efficiency, i.e. ResNet-18 \cite{he2016deep}, WideResNet-28-10 \cite{zagoruyko2016wide} and PyramidNet-110 \cite{han2017deep} on CIFAR-10 and CIFAR-100, ResNet-18 and MobileNets \cite{howard2017mobilenets} on Tiny-ImageNet.

\subsubsection{Baselines and the State-Of-The-Arts (SOTAs)}
We take the vanilla SGD and SAM \cite{foret-2020-SAM-ICLR} as baselines.
To comprehensively evaluate the performance of ARSAM, we have also chosen some SOTA methods, ESAM \cite{du-2022-ESAM-ICLR}, SAM-5 \cite{liu-2022-looksam-CVPR}, LookSAM \cite{liu-2022-looksam-CVPR}, SAF \cite{du-2022-saf-NIPS}, MESA \cite{du-2022-saf-NIPS} and AE-SAM \cite{jiang-2022-aesam-ICLR} for comparison.
These efficient methods are the follow-up works of SAM that aim to enhance efficiency.

\subsubsection{Implementation details}
On CIFAR-10 and CIFAR-100, we train all the models 200 epochs using a batch size of 128 with cutout regularization \cite{devries2017improved} and cosine learning rate decay \cite{loshchilov2016sgdr}. For Tiny-ImageNet, we also train ResNet-18 and MobileNets \cite{howard2017mobilenets} 200 epochs using a batch size of 128 with cosine learning rate decay. For ResNet-18, we use the $64 \times 64$ resolution images and transform the first convolution of the models to a $3 \times 3$ convolution. For MobileNets, we resize the original image to $224 \times 224$. $M$ was 50 for all experiments. We implement ARSAM in Pytorch and train models on a single NVIDIA GeForce RTX 3090 three times with different seed and reported the average results and standard deviation. The implementation details can be found in Appendix.

In this paper, optimization efficiency is quantified by the Average number of Images processed by the model per Second (AIS) as follows:
\begin{equation}\label{equ7}
AIS = ({D \cdot E})/{T},
\end{equation}
where $D$ represents the amount of data for one epoch of training, $T$ is the total training time in seconds, including both data loading and model optimization time, and $E$ represents the training epoch.
Additionally, we report the proportion of times the gradient of SAM was calculated during the training iterations, denoted as \%SAM. Since the runtime of the program varies across different devices, we use \%SAM and AIS for fair comparison. The same evaluation method is also used in \cite{jiang-2022-aesam-ICLR}.

\begin{table}[]
\center
\begin{tabular}{lccc}
\toprule
$\alpha$ & Accuracy & \%SAM & AIS \\ \midrule
SAM                    &     84.71    &    100\%           &     351(100\%)   \\  \midrule
0.1                   &     83.07    &    7.4\%            & \bf{620(177\%)}  \\
0.2                   &     83.63    &    16.5\%           &     581(166\%)   \\
0.3                   &     84.39    &    27.7\%           &     523(149\%)   \\
0.4                   & \bf{84.67}    &    38.2\%           &     486(139\%)   \\
0.5                   & 84.65   &    40.4\%           &     482(137\%)   \\\bottomrule
\end{tabular}
\caption{Parameter Study of $\alpha$ on CIFAR-100 datasets with WideResNet. The best accuracy and efficiency are bolded.}
\label{table_alpha}
\end{table}

\begin{table}[]
\center
\setlength{\tabcolsep}{2mm}{
\begin{tabular}{lcccc}
\toprule
   & \multicolumn{2}{c}{CIFAR-10}                                   & \multicolumn{2}{c}{CIFAR-100}                          \\ 
Methods    & Acc.  & AIS   & Acc.  & AIS \\ \midrule
SAM                & \underline{96.79}             & 1855(100\%)  
                   & \underline{81.04}            & 1866(100\%) \\
SAM-5                & 96.42             & 2741(148\%)  
                   & 80.59            & 2715(146\%) \\
ARSAM       & \bf{96.63} & 2591(140\%)  & \bf{80.92} & 2537(136\%)                                        \\
ARSAM-A     & 96.61 & 2542(137\%)  & 80.60 & 2459(134\%)                                        \\ \bottomrule
\end{tabular}}
\caption{Ablation Study of ARSAM on CIFAR-10 and CIFAR-100 with ResNet-18. The best accuracy is in bold and the second best is underlined.}
\label{table_ablation}
\end{table} 

\begin{table*}
\center
\begin{tabular}{cc|ccc|ccc}
\toprule
          & & \multicolumn{3}{c|}{\bf{CIFAR-10}}                                                        & \multicolumn{3}{c}{\bf{CIFAR-100}}                                                        \\ 
&     & Accuracy $\uparrow$ & \%SAM $\downarrow$ & AIS $\uparrow$            & Accuracy $\uparrow$ & \%SAM $\downarrow$ & AIS $\uparrow$            \\ \midrule\midrule
     \multirow{10}{*}{{\rotatebox[origin=c]{90}{\textit{ResNet-18}}}}
&SGD                & 96.23\scriptsize{$\pm$0.11}                & \textbackslash{}   & 3153(170\%) 
                   & 79.84\scriptsize{$\pm$0.45}                & \textbackslash{}   & 3181(171\%) \\ \cmidrule{2-8}
&SAM                & \bf{96.79\scriptsize{$\pm$0.03}}    & 100\%              & 1855(100\%)  
                   & \bf{81.04\scriptsize{$\pm$0.29}}    & 100\%              & 1866(100\%) \\
&SAM-5              & 96.42\scriptsize{$\pm$0.18}                & 20\%              & 2741(148\%) 
                   & 80.59\scriptsize{$\pm$0.18}                & 10\%              & 2715(146\%) \\ \cmidrule{2-8}
&LookSAM-5          & 96.47\scriptsize{$\pm$0.13}                & 20\%              & 2134(115\%) 
                   & 80.48\scriptsize{$\pm$0.24}                & 20\%              & 2115(113\%) \\
&ESAM               & 96.58\scriptsize{$\pm$0.13}                & \textbackslash{}   & 1936(104\%) 
                   & 80.47\scriptsize{$\pm$0.31}                & \textbackslash{}   & 1953(105\%) \\
&ESAM\textsuperscript{1} %\tablefootnote[1]{We report the results in \cite{du-2022-ESAM-ICLR}. But failed to reproduce them using the officially released codes in some experiments.\label{footnote_esam}}
                   & 96.56\scriptsize{$\pm$0.08}                & \textbackslash{}   & 2409(140\%) 
                   & 80.41\scriptsize{$\pm$0.10}                & \textbackslash{}   & 2423(140\%) \\
&SAF                & 96.19\scriptsize{$\pm$0.10}                & \textbackslash{}   & 2733(147\%) 
                   & 80.03\scriptsize{$\pm$0.43}                & \textbackslash{}   & 2736(147\%) \\
&SAF\textsuperscript{2} %\tablefootnote[2]{We report the results in \cite{du-2022-saf-NIPS}. For SAF, we failed to reproduce it using the official released code on CIFAR-10 and CIFAR-100, so we reproduced it ourselves following the algorithmic flow of SAF. \label{footnote_saf}}
                   & 96.37\scriptsize{$\pm$0.02}                & \textbackslash{}   & 3213(194\%) 
                   & 80.06\scriptsize{$\pm$0.05}                & \textbackslash{}   & 3248(192\%) \\
&MESA\textsuperscript{2} %\textsuperscript{\ref{footnote_saf}} 
                   & 96.24\scriptsize{$\pm$0.02}                & \textbackslash{}   & 2780(168\%) 
                   & 79.79\scriptsize{$\pm$0.09}                & \textbackslash{}   & 2793(165\%) \\
&AE-SAM\textsuperscript{3} %\tablefootnote[3]{We report the results in \cite{jiang-2022-aesam-ICLR}.\label{footnote_aesam}}             
                   & \underline{96.63\scriptsize{$\pm$0.04}}                & 50.1\%              & \textbackslash{} 
                   & 80.48\scriptsize{$\pm$0.11}                & 49.8\%              & \textbackslash{} \\
\cmidrule{2-8}
&\bf{ARSAM}          &\underline{96.63\scriptsize{$\pm$0.09}}            & 27.7\%              & 2591(140\%) 
                   &\underline{80.92\scriptsize{$\pm$0.01}}            & 23.6\%              & 2537(136\%) \\ \midrule\midrule
\multirow{10}{*}{{\rotatebox[origin=c]{90}{\textit{WideResNet-28-10}}}}
&SGD                & 96.91\scriptsize{$\pm$0.06}                & \textbackslash{}   & 680(195\%)  
                   & 82.47\scriptsize{$\pm$0.3}                & \textbackslash{}   & 685(195\%) \\ \cmidrule{2-8}
&SAM                & \bf{97.44\scriptsize{$\pm$0.04}}           & 100\%              & 349(100\%)  
                   & \bf{84.71\scriptsize{$\pm$0.21}}    & 100\%             & 351(100\%) \\
&SAM-5              & 97.16\scriptsize{$\pm$0.03}                & 20\%              & 553(158\%)  
                   & 83.06\scriptsize{$\pm$0.08}                & 20\%              & 549(156\%) \\ \cmidrule{2-8}
&LookSAM-5          & 97.13\scriptsize{$\pm$0.04}                & 20\%              & 537(154\%)  
                   & 83.52\scriptsize{$\pm$0.09}                & 20\%              & 552(157\%) \\
&ESAM               & \underline{97.41\scriptsize{$\pm$0.07}}                & \textbackslash{}   & 502(144\%)  
                   & \underline{84.71\scriptsize{$\pm$0.32}}                & \textbackslash{}   & 494(141\%) \\
&ESAM\textsuperscript{1} %\textsuperscript{\ref{footnote_esam}} 
                   & 97.29\scriptsize{$\pm$0.11}                & \textbackslash{}   & 550(139\%) 
                   & 84.51\scriptsize{$\pm$0.01}                & \textbackslash{}   & 545(139\%) \\
&SAF                & 96.96\scriptsize{$\pm$0.09}                & \textbackslash{}   & 639(183\%) 
                   & 82.57\scriptsize{$\pm$0.18}                & \textbackslash{}   & 639(182\%) \\
&SAF\textsuperscript{2} %\textsuperscript{\ref{footnote_saf}} 
                   & 97.08\scriptsize{$\pm$0.15}                & \textbackslash{}   & 727(198\%) 
                   & 83.81\scriptsize{$\pm$0.04}                & \textbackslash{}   & 729(197\%) \\
&MESA\textsuperscript{2} %\textsuperscript{\ref{footnote_saf}} 
                   & 97.16\scriptsize{$\pm$0.23}                & \textbackslash{}   & 617(168\%) 
                   & 83.59\scriptsize{$\pm$0.24}                & \textbackslash{}   & 625(169\%) \\
&AE-SAM\textsuperscript{3} %\textsuperscript{\ref{footnote_aesam}}
                   & 97.30\scriptsize{$\pm$0.10}                & 49.5\%              & \textbackslash{} 
                   & 84.51\scriptsize{$\pm$0.11}                & 49.6\%              & \textbackslash{} \\ \cmidrule{2-8}
&\bf{ARSAM}          & 97.29\scriptsize{$\pm$0.08}                & 42.5\%              & 468(134\%)  
                   &84.64\scriptsize{$\pm$0.17}          & 37.5\%              & 470(134\%)  \\ \midrule\midrule
\multirow{10}{*}{{\rotatebox[origin=c]{90}{\textit{PyramidNet-110}}}}
&SGD                & 97.14\scriptsize{$\pm$0.08}                & \textbackslash{}   & 548(195\%)  
                   & 83.38\scriptsize{$\pm$0.21}                & \textbackslash{}   & 544(196\%) \\ \cmidrule{2-8}
&SAM                &97.69\scriptsize{$\pm$0.09}                 & 100\%              & 281(100\%)  
                   &\bf{86.06\scriptsize{$\pm$0.16}}            & 100\%              & 278(100\%) \\
&SAM-5              & 97.57\scriptsize{$\pm$0.05}                & 20\%              & 447(159\%)  
                   & 84.25\scriptsize{$\pm$0.05}                & 20\%              & 457(164\%) \\ \cmidrule{2-8}
&LookSAM-5          & 97.22\scriptsize{$\pm$0.05}                & 20\%              & 339(121\%)  
                   & 83.76\scriptsize{$\pm$0.45}                & 20\%              & 350(126\%) \\
&ESAM               & 97.59\scriptsize{$\pm$0.17}                & \textbackslash{}   & 322(115\%)  
                   & 85.32\scriptsize{$\pm$0.03}                & \textbackslash{}   & 321(116\%) \\
&ESAM\textsuperscript{1} %\textsuperscript{\ref{footnote_esam}}
                   & \underline{97.81\scriptsize{$\pm$0.01}}    & \textbackslash{}   & 401(139\%) 
                   & 85.56\scriptsize{$\pm$0.05}                & \textbackslash{}   & 381(138\%) \\
&SAF                & 96.96\scriptsize{$\pm$0.05}                & \textbackslash{}   & 501(178\%) 
                   & 83.66\scriptsize{$\pm$0.34}                & \textbackslash{}   & 502(181\%) \\
&SAF\textsuperscript{2} %\textsuperscript{\ref{footnote_saf}} 
                   & 97.34\scriptsize{$\pm$0.06}                & \textbackslash{}   & 391(202\%) 
                   & 84.71\scriptsize{$\pm$0.01}                & \textbackslash{}   & 397(200\%) \\
&MESA\textsuperscript{2} %\textsuperscript{\ref{footnote_saf}}
                   & 97.46\scriptsize{$\pm$0.09}                & \textbackslash{}   & 332(171\%) 
                   & 84.73\scriptsize{$\pm$0.14}                & \textbackslash{}   & 339(171\%) \\
&AE-SAM\textsuperscript{3} %\textsuperscript{\ref{footnote_aesam}} 
                   & \bf{97.90\scriptsize{$\pm$0.05}}           & 50.3\%              & \textbackslash{} 
                   & 85.58\scriptsize{$\pm$0.10}                & 49.8\%              & \textbackslash{} \\ \cmidrule{2-8}
&\bf{ARSAM}          & 97.49\scriptsize{$\pm$0.08}                & 37.2\%              & 411(146\%) 
                   & \underline{85.67\scriptsize{$\pm$0.06}}    & 34.5\%              & 395(142\%)  \\ \bottomrule

\multicolumn{7}{l}{\small $1$ We report the results in \cite{du-2022-ESAM-ICLR}. }\\  
\multicolumn{7}{l}{\small $2$ We report the results in \cite{du-2022-saf-NIPS}. } \\                  
\multicolumn{7}{l}{\small $3$ We report the results in \cite{jiang-2022-aesam-ICLR}.}\\                   
\end{tabular}
\caption{The results of the proposed method and the comparison methods on CIFAR-10 and CIFAR-100 dataset.
The numbers in parentheses (·) indicate the ratio of corresponding method's training speed to SAM’s.
$\uparrow$ means that the larger the reported results are better, and $\downarrow$ means that the smaller the results are better.
The best accuracy is in bold and the second best is underlined.}
\label{table_result}
\end{table*}

\begin{table*}[]
\center
\setlength{\tabcolsep}{1mm}{
\begin{tabular}{lcccccccccc}
\hline
            & Head   & Shoulder & Elbow  & Wrist  & Hip    & Knee   & Ankle  & Mean   & Mean@0.1 & AIS    \\ \hline
Resnext-152 &96.248 &94.956 &88.358 &83.518 &88.298 &84.807 &80.846 & 88.662 & 33.391 & \textbackslash{} \\
Rle+ResNet50 &95.805 &94.582 &86.893 &78.346 &87.468 &80.355 &73.453 &86.024 &26.313 &\textbackslash{} \\
\hline
SimCC+HRNet+Adam       & 96.828 & \underline{95.771}   & \underline{89.654} & \underline{84.272} & \underline{88.679} & 85.151 & \underline{81.436} & 89.318 & 37.395   & 106(183\%)            \\

SimCC+HRNet+SAM   & \bf{96.965} & 95.703   & 89.637 & 84.222 & 88.402 & \underline{85.513} & \bf{81.862} & \underline{89.388} & \underline{37.721}   & 58(100\%)            \\
SimCC+HRNet+ARSAM  & \underline{96.555} & \bf{95.788}   & \bf{89.671} & \bf{84.701} & \bf{89.268} & \bf{85.614} & 81.577 & \bf{89.537} & \bf{37.799}   & 70(121\%)            \\ \hline
\end{tabular}}
\caption{Results of training SimCC on MP$\mathrm{\uppercase\expandafter{\romannumeral2}}$ with Adam, SAM and ARSAM. The best accuracy is in bold and the second best is underlined.}
\label{table_hpe}
\end{table*}

\begin{table*}[]
\centering
\begin{tabular}{cccccccccc}
\toprule
%\diagbox{CIFAR-10}{Accuracy}& \multicolumn{4}{c}{Noise rate} &\multirow{2}{*}{SAM\%} \\ \cmidrule{2-5}
 \multirow{5}{*}{{\rotatebox[origin=c]{90}{\textit{CIFAR-10}}}} &{Noise rate} & \multicolumn{2}{c}{20\%} & \multicolumn{2}{c}{40\%}& \multicolumn{2}{c}{60\%}& \multicolumn{2}{c}{80\%} \\ \cmidrule{2-10} %\cmidrule{3-4}\cmidrule{5-6}\cmidrule{7-8}\cmidrule{9-10}
& Methods   & Acc. &AIS  & Acc. &AIS & Acc. &AIS  & Acc. &AIS \\ \cmidrule{2-10}
&SGD                      &85.94    &2673(143\%)    &70.25     & 2673(143\%) &48.50  &  2575(139\%)   &28.91      & 2646(142\%)\\
&SAM                      &90.44  &  1873(100\%)    &79.71   &  1876(100\%)  &67.22   & 1858(100\%)   &{76.94}      & 1863(100\%)\\
&ARSAM                    &\bf{92.27}      & 2298(123\%) &\bf{89.08}  &  2315(123\%)   &\bf{77.67}   & 2283(123\%)   &\bf{77.93}      & 2267(122\%)\\ \midrule
 \multirow{5}{*}{{\rotatebox[origin=c]{90}{\textit{CIFAR-100}}}}&{Noise rate} & \multicolumn{2}{c}{20\%} & \multicolumn{2}{c}{40\%}& \multicolumn{2}{c}{60\%}& \multicolumn{2}{c}{80\%} \\ \cmidrule{2-10}
  & Methods   & Acc. &AIS  & Acc. &AIS & Acc. &AIS  & Acc. &AIS \\ \cmidrule{2-10}
 &SGD    &65.42    &  2537(140\%)  &48.91    &  2535(137\%) &31.46     &  2539(137\%)&12.32    & 2533(138\%) \\
  &SAM   &{69.10}   &   1815(100\%)  &55.06   & 1848(100\%)   & 35.59  &  1849(100\%)  & 9.82     &1833(100\%)\\
 &ARSAM               & \bf{69.68}    & 2237(123\%)  & \bf{55.55} & 2288(124\%)    & \bf{48.98}    & 2264(122\%) & \bf{19.18}    & 2227(122\%) \\ \bottomrule
\end{tabular}%}
\caption{Test accuracy and AIS for ResNet-18 trained on CIFAR-10 and CIFAR-100 with different rate of noisy labels. The best accuracy is in bold.}
\label{noisy_label}
\end{table*}

\begin{table}[]
\center
\setlength{\tabcolsep}{1.5mm}{
\begin{tabular}{lcccc}
\hline
Models    & Methods & Acc. $\uparrow$      &\%SAM $\downarrow$  & AIS $\uparrow$        \\ \hline
          & SGD     & 61.88\scriptsize{$\pm$0.31}& \textbackslash{} & 1456(199\%) \\
ResNet-18 & SAM     & \underline{64.48}\scriptsize{$\pm$0.15}& 100\% & 732(100\%) \\
          & ARSAM    & \bf{64.71}\scriptsize{$\pm$0.10} & 61.4\% & 892(122\%)           \\ \hline
          & SGD     & 58.12\scriptsize{$\pm$0.40}& \textbackslash{} & 547(189\%) \\
MobileNet & SAM     & \underline{58.49}\scriptsize{$\pm$0.38}& 100\% & 289(100\%) \\
          & ARSAM    & \bf{59.29}\scriptsize{$\pm$0.28} & 50.2\% &  704(122\%)  \\ \hline
\end{tabular}}
\caption{The results of SGD, SAM and ARSAM on Tiny-ImageNet. The best accuracy is in bold and the second best is underlined.}
\label{result_tiny}
\end{table}

\subsection{Parameter Studies}\label{subsec_para}
We study the effect of $\alpha$ using the WideResNet-28-10 model on the CIFAR-100 dataset. The results in Table~\ref{table_alpha} show that as $alpha$ increases, the number of times to compute PSF also increases, slowing the training speed. However, accuracy improves with more times to compute PSF.
Based on our empirical experience, we recommend setting
$\alpha$ between 0.1 and 1. For challenge datasets and models with large parameters, a higher $\alpha$ is preferable. In this paper, we set $\alpha = 0.2$ for CIFAR-10, $\alpha = 0.4$ for CIFAR-100 to achieve a 40\% improvement in optimization speed compared to SAM, and $\alpha = 0.8$ for the more challenging Tiny-ImageNet dataset.

\subsection{Comparison to SOTAs}
We trained ResNet-18, WideResNet-28-10, and PyramidNet-110 using different optimizer on CIFAR-10 and CIFAR-100, with results shown in Table~\ref{table_result}. For Tiny-ImageNet, we trained ResNet-18 and MobileNets using SGD, SAM, and ARSAM, with results in Table~\ref{result_tiny}.

\textbf{Accuracy.} From Table~\ref{table_result}, ARSAM achieves significantly higher accuracy than SGD and SAM-5, and its accuracy is comparable to SAM. Compared to LookSAM, ESAM, and AE-SAM, ARSAM outperforms in most cases, demonstrating its ability to preserve generalization during training. Because ARSAM calculates the PSF in more appropriate optimization iterations, it promotes generalization. ARSAM reuses the PSF also guarantees the generalization ability of the model. ARSAM outperforms SAF and MESA because these methods optimize sharpness over several iterations, which may overlook local sharpness, as multiple local sharpness often occurs during multiple optimization iterations~\cite{zhang-2023-gradient-CVPR}. 
For Tiny-ImageNet, ARSAM achieves results comparable to SAM on ResNet-18 and MobileNets, showing its versatility across datasets and models.

\textbf{Efficiency.} In Table~\ref{table_result}, SAM’s optimization speed is about half that of SGD for WideResNet-28-10 and PyramidNet-110, but slightly faster than half for ResNet-18 due to data loading time being significant for small model but negligible for larger model. ARSAM achieves about 40\% faster training than SAM while maintaining  comparable accuracy. SAM-5 enhance optimization speed but suffer a significant decrease in accuracy. In some cases, ARSAM is faster than LookSAM and ESAM. Because LookSAM includes extra computations like gradient projection, and ESAM still requires two gradient computations per optimization step. SAF sacrifices memory for optimization efficiency, achieving nearly the same speed as SGD. MESA accelerates training by utilizing an EMA model and eliminating the need for backward when updating the EMA model. Although ARSAM may not match the speed of SAF and MESA, it excels at preserving the model's generalization ability. ARSAM also achieves comparable performance to AE-SAM with fewer SAM updates. For Tiny-ImageNet, the training efficiency of ARSAM is higher than that of SAM.

\subsection{Ablation Study}
To better understand the impact of gradient reuse on performance and efficiency, we introduce a variant of ARSAM that does not utilize gradient reuse, referred to as ARSAM-A. The results are shown in Table~\ref{table_ablation}. Comparing ARSAM with ARSAM-A, we observe that the gradient reuse strategy can prevent the model's generalization ability without increasing the computational cost.

\subsection{Robustness to Label Noise}
Previous research has shown that SAM is robust to label noise. This subsection explores the effectiveness of ARSAM in the classic noisy-label setting for CIFAR-10 and CIFAR-100. We simulate symmetric label noise by randomly flipping labels \cite{huang2019o2u}. In these experiments, we set $\alpha=0.3$ for CIFAR-10 and $\alpha=0.4$ for CIFAR-100. 

As shown in Table \ref{noisy_label}, for CIFAR-10, models optimized with SGD experience a sharp accuracy drop, falling to 28.91\% at 80\% label noise. In contrast, SAM and ARSAM models maintain around 70\% accuracy even at 80\% noise. A similar trend is observed for CIFAR-100, though SAM performs worse than SGD at 80\% noise, likely due to SAM's difficulty in converging under high label noise, as noted in \cite{foret-2020-SAM-ICLR}. However, ARSAM consistently outperforms both SGD and SAM, demonstrating superior robustness in high-noise conditions. This can be attributed to ARSAM's predominant use of SGD in the early stages of training to efficiently identify local minima and learn the clean samples. In the later stages, ARSAM transitions to leveraging SAM to find flatter minima, thereby enhancing generalization. Additionally, ARSAM achieves approximately a 25\% speedup compared to SAM.

\subsection{Application to Semantic Segmentation}
We conduct semantic segmentation experiments on the PASCAL VOC 2012 dataset \cite{everingham2010pascal}, which includes 20 object categories and one background class. PSPNet \cite{zhao2017pyramid} and DeepLabv3+ \cite{chen2018encoder} are selected as representative models. Following \cite{zhao2017pyramid}, we adopt the augmented annotations from \cite{hariharan2011semantic}, resulting in 10,582 training, 1,449 validation, and 1,456 test images.

Table~\ref{table_ss} shows the semantic segmentation results on VOC 2012 using PSPNet and DeepLabv3+ with SGD, SAM, and ARSAM. On PSPNet, SAM achieves the highest accuracy of 79.20\%, while ARSAM closely follows with 79.10\%, outperforming SGD (78.89\%). On DeepLabv3+, ARSAM achieves the best accuracy of 77.56\%. ARSAM improves the optimization speed by approximately 32\% on PSPNet and 31\% on DeepLabv3+ compared to SAM, while still maintaining competitive accuracy.

\begin{table}[t]
\centering
\setlength{\tabcolsep}{0.5mm}{
\begin{tabular*}{\linewidth}{l@{\extracolsep{\fill}}cccc}
\toprule
\multirow{2}{*}{Methods}           & \multicolumn{2}{c}{PSPNet}  & \multicolumn{2}{c}{DeepLabv3+} \\ \cmidrule{2-3}\cmidrule{4-5}
    & Acc.      & AIS     & Acc.    & AIS   \\ \midrule
SGD    & 78.89     & 37(195\%)    & 77.12   & 152(190\%)          \\
SAM    & \bf{79.20}     & 19(100\%)     & \underline{77.25}   & 80(100\%)           \\
ARSAM   & \underline{79.10}     & 25(132\%)    & \bf{77.56}   & 105(131\%)          \\ \bottomrule
\end{tabular*}}
\caption{Results of semantic segmentation with SGD, SAM, and ARSAM on the VOC 2012. The best accuracy is in bold and the second best is underlined.}
\label{table_ss}
\end{table}

\subsection{Application to Human Pose Estimation}
2D Human Pose Estimation (HPE) focuses on localizing body joints within a single image. To evaluate the generality of our methods, we apply SAM and ARSAM to human pose estimation. SimCC \cite{li2022simcc} is a SOTA method for HPE. The key idea of SimCC is to treat HPE as two classification tasks. 

We employ SAM and ARSAM as optimization methods for the SimCC, using Adam as the base optimizer for both. We conducted experiments on the MP$\mathrm{\uppercase\expandafter{\romannumeral2}}$ dataset \cite{andriluka14cvpr}, which contains 40k person samples with 16 joints labels. Following the evaluation procedure outlined in \cite{li2022simcc}, we performed experiments at an image resolution of $256 \times 256$, with a batch size of 64. For comparison, we also report the results of ResNeXt-152 \cite{xie2017aggregated} and RLE+ResNet-50 \cite{li2021human}, sourced from the MMPose project \cite{mmpose2020}.
As shown in Table \ref{table_hpe}, ARSAM achieves performance comparable to SAM while significantly reducing training time. This demonstrates that ARSAM is simple, versatile, and effective. Moreover, ARSAM remains effective when using different base optimizers, such as Adam.

\subsection{Application to Quantization-Aware Training}
Neural quantization reduces the computational workload and memory requirements of deep neural networks by compressing the precision of weights and activation values into low-bit formats \cite{esser2020learned,wei2021qdrop,nagel2022overcoming}. To demonstrate ARSAM's broader applicability, we use it as the optimizer for QAT \cite{jacob-2018-quantization}. 

We used SGD, SAM, and ARSAM to quantize the parameters of ResNet-18 and MobileNets to W4A4 on the CIFAR-10 dataset. The results, shown in Table~\ref{table_LSQ}, indicate that ARSAM achieves about 50\% faster training speeds compared to SAM while maintaining comparable performance. This demonstrates that ARSAM is versatile and satisfies the practical requirements of various applications.

\begin{table}[]
\center
\resizebox{\linewidth}{!}{
\begin{tabular}{lcccc}
\toprule
           & \multicolumn{2}{c}{ResNet-18} & \multicolumn{2}{c}{MobileNets} \\ 
Methods    & Acc.      & AIS     & Acc.    & AIS   \\ \hline
Full prec. & 88.72     & \textbackslash{}  & 85.81      & \textbackslash{}           \\
LSQ+SGD    & 88.86     & 1515(174\%)    & 84.04   & 1244(153\%)          \\
LSQ+SAM    & \bf{89.75}     & 870(100\%)     & \underline{84.72}   & 816(100\%)           \\
LSQ+ARSAM   & \underline{89.63}     & 1386(159\%)    & \bf{84.77}   & 1199(147\%)          \\ \bottomrule
\end{tabular}
}
\caption{Results of optimizing LSQ using ARSAM on the Cifar-10 dataset. The best accuracy is in bold and the second best is underlined.}
\label{table_LSQ}
\end{table}

\section{Conclusions}
In this paper, we reveal that the gradient of SAM can be decomposed into the gradient of SGD and the PSF. We propose an adaptive sampling-reusing-mixing policy for the decomposed gradients, aimed at reducing the computational cost of the PSF and approximating the true PSF. Using this policy, we introduce ARSAM, an efficient optimization method that adaptively adjusts the gradient sampling rate of the PSF to improve optimization efficiency. It reuses the PSF in non-sampling iterations, thereby maintaining the model's generalization ability. We theoretically and empirically analyzed the convergence of ARSAM. Experimental results demonstrate that ARSAM achieves similar generalization performance to SAM while significantly improving optimization speed. Furthermore, we applied ARSAM to human pose estimation and model quantization tasks. The results confirm that ARSAM enhances optimization speed without compromising performance, showcasing its broad applicability. 

In some cases, ARSAM achieves even higher accuracy than SAM. This indicates that the reuse of the PSF could reach the a flatter region than that of SAM. Investigating the underlying reasons for this phenomenon remains a future research direction. Besides, the autoregressive approach faces challenges in conveniently controlling the acceleration ratio of ARSAM relative to SAM. In the future, we aim to explore improved methods for modeling the relationship between the importance of the PSF and the sampling ratio to address this limitation.

\bibliographystyle{named}
\bibliography{ijcai25}

\begin{thebibliography}{}

\bibitem[\protect\citeauthoryear{Andriluka \bgroup \em et al.\egroup }{2014}]{andriluka14cvpr}
Mykhaylo Andriluka, Leonid Pishchulin, Peter Gehler, and Bernt Schiele.
\newblock 2d human pose estimation: New benchmark and state of the art analysis.
\newblock In {\em IEEE Conference on Computer Vision and Pattern Recognition (CVPR)}, June 2014.

\bibitem[\protect\citeauthoryear{Andriushchenko and Flammarion}{2022}]{andriushchenko-2022-towards_understanding-ICML}
Maksym Andriushchenko and Nicolas Flammarion.
\newblock Towards understanding sharpness-aware minimization.
\newblock In {\em Proceedings of the International Conference on Machine Learning (ICML)}, pages 639--668. PMLR, 2022.

\bibitem[\protect\citeauthoryear{Chaudhari \bgroup \em et al.\egroup }{2019}]{chaudhari-2019-Entropy_sgd-IOP}
Pratik Chaudhari, Anna Choromanska, Stefano Soatto, Yann LeCun, Carlo Baldassi, Christian Borgs, Jennifer Chayes, Levent Sagun, and Riccardo Zecchina.
\newblock Entropy-sgd: Biasing gradient descent into wide valleys.
\newblock {\em Journal of Statistical Mechanics: Theory and Experiment}, 2019(12):124018, 2019.

\bibitem[\protect\citeauthoryear{Chen \bgroup \em et al.\egroup }{2018}]{chen2018encoder}
Liang-Chieh Chen, Yukun Zhu, George Papandreou, Florian Schroff, and Hartwig Adam.
\newblock Encoder-decoder with atrous separable convolution for semantic image segmentation.
\newblock In {\em Proceedings of the European conference on computer vision (ECCV)}, pages 801--818, 2018.

\bibitem[\protect\citeauthoryear{Contributors}{2020}]{mmpose2020}
MMPose Contributors.
\newblock Openmmlab pose estimation toolbox and benchmark.
\newblock \url{https://github.com/open-mmlab/mmpose}, 2020.

\bibitem[\protect\citeauthoryear{Dai \bgroup \em et al.\egroup }{2023}]{FedGAMMA}
Rong Dai, Xun Yang, Yan Sun, Li~Shen, Xinmei Tian, Meng Wang, and Yongdong Zhang.
\newblock Fedgamma: Federated learning with global sharpness-aware minimization.
\newblock {\em IEEE Transactions on Neural Networks and Learning Systems}, pages 1--14, 2023.
\newblock doi:{10.1109/TNNLS.2023.3304453}.

\bibitem[\protect\citeauthoryear{Deng \bgroup \em et al.\egroup }{2024}]{Deng2024AsymptoticUS}
Jiaxin Deng, Junbiao Pang, and Baochang Zhang.
\newblock Asymptotic unbiased sample sampling to speed up sharpness-aware minimization.
\newblock {\em ArXiv}, abs/2406.08001, 2024.

\bibitem[\protect\citeauthoryear{DeVries and Taylor}{2017}]{devries2017improved}
Terrance DeVries and Graham~W Taylor.
\newblock Improved regularization of convolutional neural networks with cutout.
\newblock {\em arXiv preprint arXiv:1708.04552}, 2017.

\bibitem[\protect\citeauthoryear{Dinh \bgroup \em et al.\egroup }{2017}]{dinh-2017-sharp_minima-ICML}
Laurent Dinh, Razvan Pascanu, Samy Bengio, and Yoshua Bengio.
\newblock Sharp minima can generalize for deep nets.
\newblock In {\em Proceedings of the International Conference on Machine Learning (ICML)}, pages 1019--1028. PMLR, 2017.

\bibitem[\protect\citeauthoryear{Dong \bgroup \em et al.\egroup }{2024}]{dong2024implicit}
Mingrong Dong, Yixuan Yang, Kai Zeng, Qingwang Wang, and Tao Shen.
\newblock Implicit sharpness-aware minimization for domain generalization.
\newblock {\em Remote Sensing}, 16(16):2877, 2024.

\bibitem[\protect\citeauthoryear{Du \bgroup \em et al.\egroup }{2022a}]{du-2022-ESAM-ICLR}
Jiawei Du, Hanshu Yan, Jiashi Feng, Joey~Tianyi Zhou, Liangli Zhen, Rick Siow~Mong Goh, and Vincent~YF Tan.
\newblock Efficient sharpness-aware minimization for improved training of neural networks.
\newblock In {\em Proceedings of the International Conference on Learning Representations (ICLR)}, 2022.

\bibitem[\protect\citeauthoryear{Du \bgroup \em et al.\egroup }{2022b}]{du-2022-saf-NIPS}
Jiawei Du, Daquan Zhou, Jiashi Feng, Vincent Tan, and Joey~Tianyi Zhou.
\newblock Sharpness-aware training for free.
\newblock {\em Advances in Neural Information Processing Systems (NeurIPS)}, 35:23439--23451, 2022.

\bibitem[\protect\citeauthoryear{Esser \bgroup \em et al.\egroup }{2020}]{esser2020learned}
Steven~K Esser, Jeffrey~L McKinstry, Deepika Bablani, Rathinakumar Appuswamy, and Dharmendra~S Modha.
\newblock Learned step size quantization.
\newblock In {\em Proceedings of the International Conference on Learning Representations (ICLR)}, 2020.

\bibitem[\protect\citeauthoryear{Everingham \bgroup \em et al.\egroup }{2010}]{everingham2010pascal}
Mark Everingham, Luc Van~Gool, Christopher~KI Williams, John Winn, and Andrew Zisserman.
\newblock The pascal visual object classes (voc) challenge.
\newblock {\em International journal of computer vision}, 88:303--338, 2010.

\bibitem[\protect\citeauthoryear{Foret \bgroup \em et al.\egroup }{2021}]{foret-2020-SAM-ICLR}
Pierre Foret, Ariel Kleiner, Hossein Mobahi, and Behnam Neyshabur.
\newblock Sharpness-aware minimization for efficiently improving generalization.
\newblock In {\em Proceedings of the International Conference on Learning Representations (ICLR)}, 2021.

\bibitem[\protect\citeauthoryear{Han \bgroup \em et al.\egroup }{2017}]{han2017deep}
Dongyoon Han, Jiwhan Kim, and Junmo Kim.
\newblock Deep pyramidal residual networks.
\newblock In {\em Proceedings of the Conference on Computer Vision and Pattern Recognition (CVPR)}, pages 5927--5935, 2017.

\bibitem[\protect\citeauthoryear{Hariharan \bgroup \em et al.\egroup }{2011}]{hariharan2011semantic}
Bharath Hariharan, Pablo Arbel{\'a}ez, Lubomir Bourdev, Subhransu Maji, and Jitendra Malik.
\newblock Semantic contours from inverse detectors.
\newblock In {\em 2011 international conference on computer vision}, pages 991--998. IEEE, 2011.

\bibitem[\protect\citeauthoryear{He \bgroup \em et al.\egroup }{2016}]{he2016deep}
Kaiming He, Xiangyu Zhang, Shaoqing Ren, and Jian Sun.
\newblock Deep residual learning for image recognition.
\newblock In {\em Proceedings of the Conference on Computer Vision and Pattern Recognition (CVPR)}, pages 770--778, 2016.

\bibitem[\protect\citeauthoryear{Howard \bgroup \em et al.\egroup }{2017}]{howard2017mobilenets}
Andrew~G Howard, Menglong Zhu, Bo~Chen, Dmitry Kalenichenko, Weijun Wang, Tobias Weyand, Marco Andreetto, and Hartwig Adam.
\newblock Mobilenets: Efficient convolutional neural networks for mobile vision applications.
\newblock {\em arXiv preprint arXiv:1704.04861}, 2017.

\bibitem[\protect\citeauthoryear{Huang \bgroup \em et al.\egroup }{2019}]{huang2019o2u}
Jinchi Huang, Lie Qu, Rongfei Jia, and Binqiang Zhao.
\newblock O2u-net: A simple noisy label detection approach for deep neural networks.
\newblock In {\em Proceedings of the IEEE/CVF international conference on computer vision (ICCV)}, pages 3326--3334, 2019.

\bibitem[\protect\citeauthoryear{Izmailov \bgroup \em et al.\egroup }{2018}]{izmailov-2018-SWA-UAI}
Pavel Izmailov, Dmitrii Podoprikhin, Timur Garipov, Dmitry Vetrov, and Andrew~Gordon Wilson.
\newblock Averaging weights leads to wider optima and better generalization.
\newblock In {\em Proceedings of the Conference on Uncertainty in Artificial Intelligence}, pages 876--885, 2018.

\bibitem[\protect\citeauthoryear{Jacob \bgroup \em et al.\egroup }{2018}]{jacob-2018-quantization}
Benoit Jacob, Skirmantas Kligys, Bo~Chen, Menglong Zhu, Matthew Tang, Andrew Howard, Hartwig Adam, and Dmitry Kalenichenko.
\newblock Quantization and training of neural networks for efficient integer-arithmetic-only inference.
\newblock In {\em Proceedings of the Conference on Computer Vision and Pattern Recognition (CVPR)}, pages 2704--2713, 2018.

\bibitem[\protect\citeauthoryear{Jiang \bgroup \em et al.\egroup }{2019}]{jiang-2019-fantastic-ICLR}
Yiding Jiang, Behnam Neyshabur, Hossein Mobahi, Dilip Krishnan, and Samy Bengio.
\newblock Fantastic generalization measures and where to find them.
\newblock {\em arXiv preprint arXiv:1912.02178}, 2019.

\bibitem[\protect\citeauthoryear{Jiang \bgroup \em et al.\egroup }{2023}]{jiang-2022-aesam-ICLR}
Weisen Jiang, Hansi Yang, Yu~Zhang, and James Kwok.
\newblock An adaptive policy to employ sharpness-aware minimization.
\newblock In {\em Proceedings of the International Conference on Learning Representations (ICLR)}, 2023.

\bibitem[\protect\citeauthoryear{Keskar \bgroup \em et al.\egroup }{2017}]{keskar-2016-large_batch-ICLR}
Nitish~Shirish Keskar, Dheevatsa Mudigere, Jorge Nocedal, Mikhail Smelyanskiy, and Ping Tak~Peter Tang.
\newblock On large-batch training for deep learning: Generalization gap and sharp minima.
\newblock In {\em Proceedings of the International Conference on Learning Representations (ICLR)}, 2017.

\bibitem[\protect\citeauthoryear{Krizhevsky \bgroup \em et al.\egroup }{2009}]{krizhevsky2009learning}
Alex Krizhevsky, Geoffrey Hinton, et~al.
\newblock Learning multiple layers of features from tiny images.
\newblock 2009.

\bibitem[\protect\citeauthoryear{Kwon \bgroup \em et al.\egroup }{2021}]{kwon-2021-asam-ICML}
Jungmin Kwon, Jeongseop Kim, Hyunseo Park, and In~Kwon Choi.
\newblock Asam: Adaptive sharpness-aware minimization for scale-invariant learning of deep neural networks.
\newblock In {\em Proceedings of the International Conference on Machine Learning (ICML)}, pages 5905--5914. PMLR, 2021.

\bibitem[\protect\citeauthoryear{Le and Yang}{2015}]{le-2015-tiny}
Ya~Le and Xuan Yang.
\newblock Tiny imagenet visual recognition challenge.
\newblock {\em CS 231N}, 7(7):3, 2015.

\bibitem[\protect\citeauthoryear{Li \bgroup \em et al.\egroup }{2018}]{li-2018-visualizing-NIPS}
Hao Li, Zheng Xu, Gavin Taylor, Christoph Studer, and Tom Goldstein.
\newblock Visualizing the loss landscape of neural nets.
\newblock {\em Advances in Neural Information Processing Systems (NeurIPS)}, pages 6391--6401, 2018.

\bibitem[\protect\citeauthoryear{Li \bgroup \em et al.\egroup }{2021}]{li2021human}
Jiefeng Li, Siyuan Bian, Ailing Zeng, Can Wang, Bo~Pang, Wentao Liu, and Cewu Lu.
\newblock Human pose regression with residual log-likelihood estimation.
\newblock In {\em Proceedings of the IEEE/CVF International Conference on Computer Vision}, pages 11025--11034, 2021.

\bibitem[\protect\citeauthoryear{Li \bgroup \em et al.\egroup }{2022}]{li2022simcc}
Yanjie Li, Sen Yang, Peidong Liu, Shoukui Zhang, Yunxiao Wang, Zhicheng Wang, Wankou Yang, and Shu-Tao Xia.
\newblock Simcc: A simple coordinate classification perspective for human pose estimation.
\newblock In {\em European Conference on Computer Vision}, pages 89--106. Springer, 2022.

\bibitem[\protect\citeauthoryear{Liu \bgroup \em et al.\egroup }{2020}]{liu-2020-loss-NIPS}
Chen Liu, Mathieu Salzmann, Tao Lin, Ryota Tomioka, and Sabine S{\"u}sstrunk.
\newblock On the loss landscape of adversarial training: Identifying challenges and how to overcome them.
\newblock {\em Advances in Neural Information Processing Systems (NeurIPS)}, 33:21476--21487, 2020.

\bibitem[\protect\citeauthoryear{Liu \bgroup \em et al.\egroup }{2022}]{liu-2022-looksam-CVPR}
Yong Liu, Siqi Mai, Xiangning Chen, Cho-Jui Hsieh, and Yang You.
\newblock Towards efficient and scalable sharpness-aware minimization.
\newblock In {\em Proceedings of the Conference on Computer Vision and Pattern Recognition (CVPR)}, pages 12360--12370, 2022.

\bibitem[\protect\citeauthoryear{Loshchilov and Hutter}{2016}]{loshchilov2016sgdr}
Ilya Loshchilov and Frank Hutter.
\newblock Sgdr: Stochastic gradient descent with warm restarts.
\newblock {\em arXiv preprint arXiv:1608.03983}, 2016.

\bibitem[\protect\citeauthoryear{Nagel \bgroup \em et al.\egroup }{2022}]{nagel2022overcoming}
Markus Nagel, Marios Fournarakis, Yelysei Bondarenko, and Tijmen Blankevoort.
\newblock Overcoming oscillations in quantization-aware training.
\newblock In {\em Proceedings of the International Conference on Machine Learning (ICML)}, pages 16318--16330. PMLR, 2022.

\bibitem[\protect\citeauthoryear{Sun \bgroup \em et al.\egroup }{2021}]{sun-2021-exploring-AAAI}
Xu~Sun, Zhiyuan Zhang, Xuancheng Ren, Ruixuan Luo, and Liangyou Li.
\newblock Exploring the vulnerability of deep neural networks: A study of parameter corruption.
\newblock In {\em Proceedings of the AAAI Conference on Artificial Intelligence (AAAI)}, volume~35, pages 11648--11656, 2021.

\bibitem[\protect\citeauthoryear{Wei \bgroup \em et al.\egroup }{2021}]{wei2021qdrop}
Xiuying Wei, Ruihao Gong, Yuhang Li, Xianglong Liu, and Fengwei Yu.
\newblock Qdrop: Randomly dropping quantization for extremely low-bit post-training quantization.
\newblock In {\em Proceedings of the International Conference on Learning Representations (ICLR)}, 2021.

\bibitem[\protect\citeauthoryear{Xie \bgroup \em et al.\egroup }{2017}]{xie2017aggregated}
Saining Xie, Ross Girshick, Piotr Doll{\'a}r, Zhuowen Tu, and Kaiming He.
\newblock Aggregated residual transformations for deep neural networks.
\newblock In {\em Proceedings of the IEEE conference on computer vision and pattern recognition}, pages 1492--1500, 2017.

\bibitem[\protect\citeauthoryear{Xie \bgroup \em et al.\egroup }{2024}]{xie2024adaptive}
Tianci Xie, Tao Li, and Ruoxue Wu.
\newblock Adaptive sharpness-aware minimization for adversarial domain generalization.
\newblock {\em IEEE Transactions on Computational Social Systems}, 2024.

\bibitem[\protect\citeauthoryear{Yue \bgroup \em et al.\egroup }{2023}]{Yue2023SALR}
Xubo Yue, Maher Nouiehed, and Raed~Al Kontar.
\newblock Salr: Sharpness-aware learning rate scheduler for improved generalization.
\newblock {\em IEEE Transactions on Neural Networks and Learning Systems}, pages 1--10, 2023.
\newblock doi:{10.1109/TNNLS.2023.3263393}.

\bibitem[\protect\citeauthoryear{Zagoruyko and Komodakis}{2016}]{zagoruyko2016wide}
Sergey Zagoruyko and Nikos Komodakis.
\newblock Wide residual networks.
\newblock {\em arXiv preprint arXiv:1605.07146}, 2016.

\bibitem[\protect\citeauthoryear{Zhang \bgroup \em et al.\egroup }{2023}]{zhang-2023-gradient-CVPR}
Xingxuan Zhang, Renzhe Xu, Han Yu, Hao Zou, and Peng Cui.
\newblock Gradient norm aware minimization seeks first-order flatness and improves generalization.
\newblock In {\em Proceedings of the Conference on Computer Vision and Pattern Recognition (CVPR)}, pages 20247--20257, 2023.

\bibitem[\protect\citeauthoryear{Zhao \bgroup \em et al.\egroup }{2017}]{zhao2017pyramid}
Hengshuang Zhao, Jianping Shi, Xiaojuan Qi, Xiaogang Wang, and Jiaya Jia.
\newblock Pyramid scene parsing network.
\newblock In {\em Proceedings of the IEEE conference on computer vision and pattern recognition}, pages 2881--2890, 2017.

\bibitem[\protect\citeauthoryear{Zhao \bgroup \em et al.\egroup }{2022}]{zhao-2022-penalizing-ICML}
Yang Zhao, Hao Zhang, and Xiuyuan Hu.
\newblock Penalizing gradient norm for efficiently improving generalization in deep learning.
\newblock In {\em Proceedings of the International Conference on Machine Learning (ICML)}, pages 26982--26992. PMLR, 2022.

\bibitem[\protect\citeauthoryear{Zhou \bgroup \em et al.\egroup }{2023}]{zhou2023imbsam}
Yixuan Zhou, Yi~Qu, Xing Xu, and Hengtao Shen.
\newblock Imbsam: A closer look at sharpness-aware minimization in class-imbalanced recognition.
\newblock In {\em Proceedings of the IEEE/CVF International Conference on Computer Vision}, pages 11345--11355, 2023.

\bibitem[\protect\citeauthoryear{Zhuang \bgroup \em et al.\egroup }{2022}]{zhuang-2022-GSAM-ICLR}
Juntang Zhuang, Boqing Gong, Liangzhe Yuan, Yin Cui, Hartwig Adam, Nicha Dvornek, Sekhar Tatikonda, James Duncan, and Ting Liu.
\newblock Surrogate gap minimization improves sharpness-aware training.
\newblock In {\em Proceedings of the International Conference on Learning Representations (ICLR)}, 2022.

\end{thebibliography}

\end{document}